\documentclass[11pt,a4paper]{article}
\usepackage{authblk}
\usepackage[hyperref]{acl2020}
\usepackage{graphicx}
\usepackage{multirow}
\usepackage{times}
\usepackage{latexsym}

\usepackage{microtype}

\aclfinalcopy 

\title{Rethinking Self-Attention: \\ Towards Interpretability in Neural Parsing}

\author[1]{\bf Khalil Mrini}
\author[2]{\bf Franck Dernoncourt}
\author[2]{\bf Quan Tran}
\author[2]{\\ \bf Trung Bui}
\author[2]{\bf Walter Chang}
\author[1]{\bf Ndapa Nakashole}
\affil[1]{
University of California, San Diego,
La Jolla, CA 92093 \protect\\
\small{\texttt{khalil@ucsd.edu, nnakashole@eng.ucsd.edu}}}
\affil[2]{Adobe Research,
San Jose, CA 95110 \protect\\
\small{\texttt{\{franck.dernoncourt, qtran, bui, wachang\}@adobe.com}}}

\date{}

\begin{document}
\maketitle
\begin{abstract}
Attention mechanisms have improved the performance of NLP tasks while allowing models to remain explainable.
Self-attention is currently widely used, however interpretability is difficult due to the numerous attention distributions. Recent work has shown that model representations can benefit from label-specific information, while facilitating interpretation of predictions. We introduce the Label Attention Layer: a new form of self-attention where attention heads represent labels. We test our novel layer by running constituency and dependency parsing experiments and show our new model obtains new state-of-the-art results for both tasks on both the Penn Treebank (PTB) and Chinese Treebank.
Additionally, our model requires fewer self-attention layers compared to existing work.
Finally, we find that the Label Attention heads learn relations between syntactic categories and show pathways to analyze errors.
\end{abstract}

\section{Introduction}

\begin{figure}
    \centering
    \includegraphics[width=220pt]{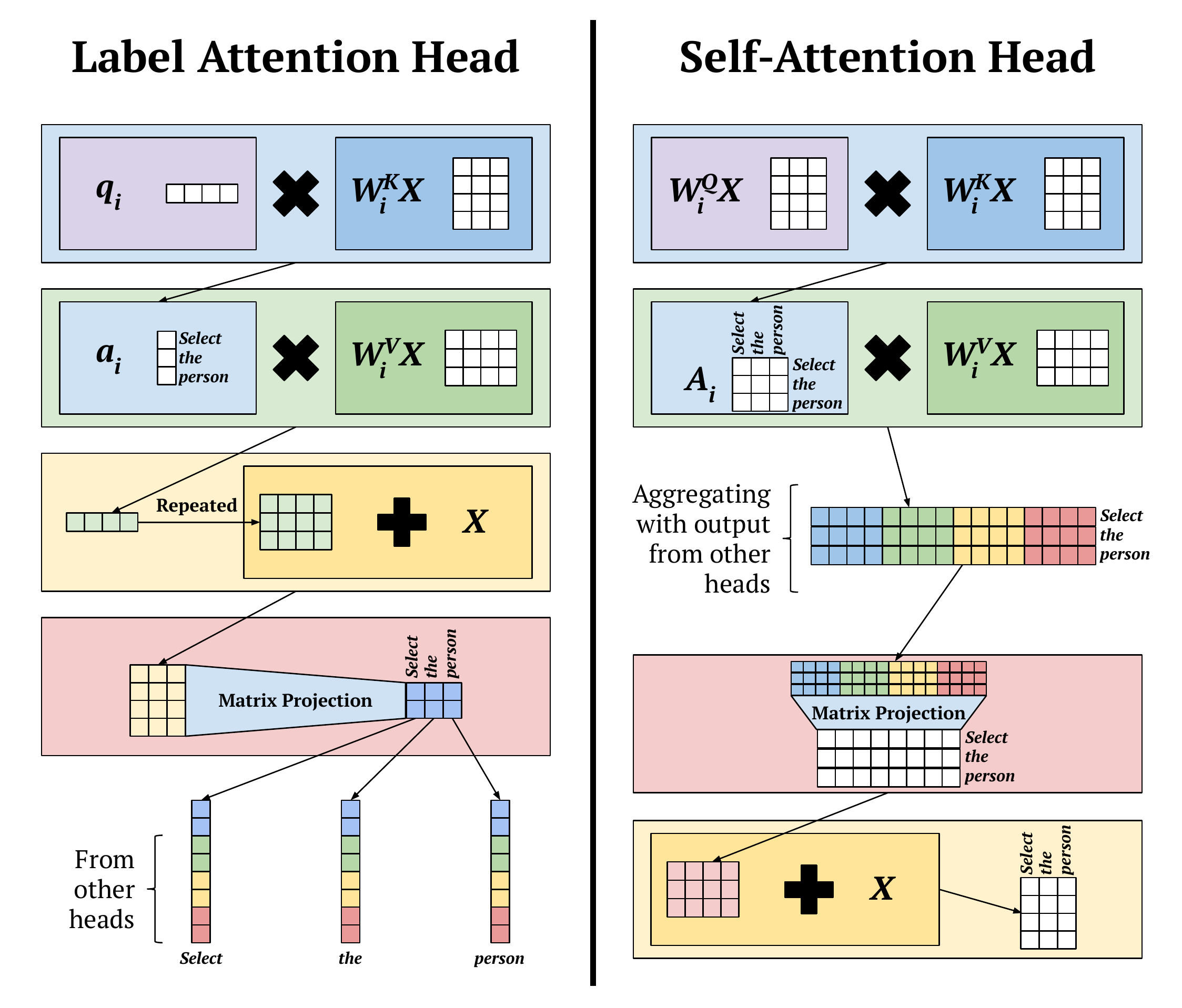}
    \caption{Comparison of the attention head architectures of our proposed Label Attention Layer and a Self-Attention Layer \cite{vaswani2017attention}. The matrix $\mathbf{X}$ represents the input sentence ``\textit{Select the person}''.}
    \label{lalc}
\end{figure}

Attention mechanisms \cite{bahdanau2014neural, luong2015effective} provide arguably explainable attention distributions that can help to interpret predictions. For example, for their machine translation predictions, \citet{bahdanau2014neural} show a heat map of attention weights from source language words to target language words. Similarly, in transformer architectures \cite{vaswani2017attention}, a self-attention head produces attention distributions from the input words to the same input words, as shown in the second row on the right side of Figure~\ref{lalc}. However, self-attention mechanisms have multiple heads, making the combined outputs difficult to interpret.

Recent work in multi-label text classification \cite{xiao2019label} and sequence labeling \cite{cui2019hierarchically} shows the efficiency and interpretability of label-specific representations. 
We introduce the Label Attention Layer: a modified version of self-attention, where each classification label corresponds to one or more attention heads. We project the output at the attention head level, rather than after aggregating all outputs, to preserve the source of head-specific information, thus allowing us to match labels to heads.

To test our proposed Label Attention Layer, we build upon the parser of \citet{zhou2019head} and establish a new state of the art for both constituency and dependency parsing, in both English and Chinese. We also release our pre-trained parsers, as well as our code to encourage experiments with 
the Label Attention Layer
\footnote{Available at: \href{http://www.github.com/KhalilMrini/LAL-Parser}{GitHub.com/KhalilMrini/LAL-Parser}}.


\begin{figure*}
    \centering 
    \includegraphics[width=420pt]{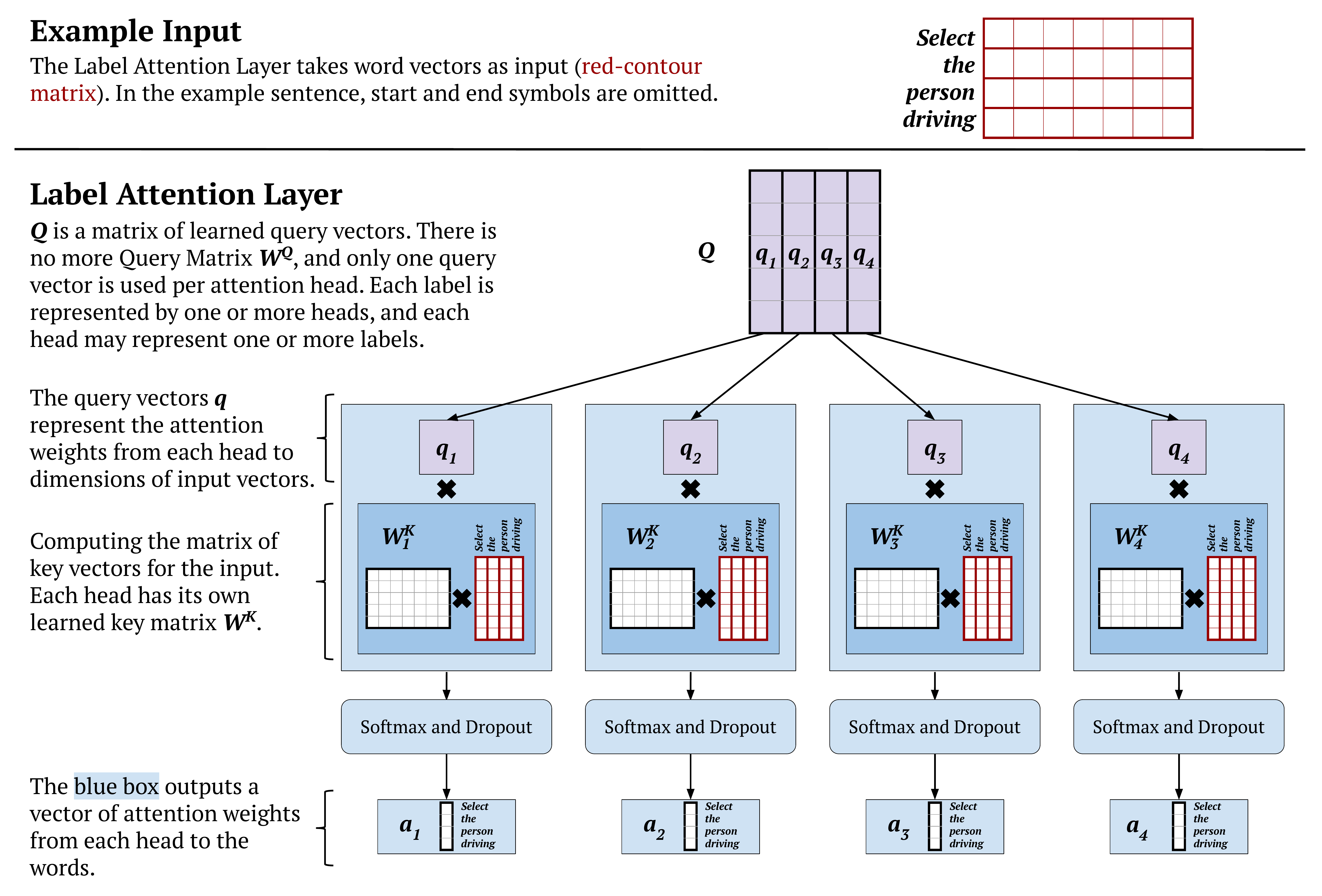}
    \caption{The architecture of the top of our proposed Label Attention Layer. In this figure, the example input sentence is ``\textit{Select the person driving}''.}
    \label{lal4}
\end{figure*}

\section{Label Attention Layer}
\label{section_lal}

The self-attention mechanism of \citet{vaswani2017attention} propagates information between the words of a sentence. Each resulting word representation contains its own attention-weighted view of the sentence. We hypothesize that a word representation can be enhanced by including each label's attention-weighted view of the sentence, on top of the information obtained from self-attention.
 
The Label Attention Layer (LAL) is a novel, modified form of self-attention, where only one query vector is needed per attention head. Each classification label is represented by one or more attention heads, and this allows the model to learn label-specific views of the input sentence. Figure \ref{lalc} shows a high-level comparison between our Label Attention Layer and self-attention.

We explain the architecture and intuition behind our proposed \textit{Label Attention Layer} through the example application of parsing.

Figure \ref{lal4} shows one of the main differences between our Label Attention mechanism and self-attention: the absence of the Query matrix $\mathbf{W^{Q}}$. Instead, we have a learned matrix $\mathbf{Q}$ of query vectors representing each head. More formally, for the attention head $i$ and an input matrix $\mathbf{X}$ of word vectors, we compute the corresponding attention weights vector $\mathbf{a}_i$ as follows:

\begin{equation}
    \mathbf{a}_i = \textup{softmax}\left(\frac{\mathbf{q}_i * \mathbf{K}_i}{\sqrt{d}}\right)
    \label{eq1}
\end{equation}

\noindent where $d$ is the dimension of query and key vectors, $\mathbf{K}_i$ is the matrix of key vectors. Given a learned head-specific key matrix $\mathbf{W}^K_{i}$, we compute $\mathbf{K}_i$ as:

\begin{equation}
    \mathbf{K}_i = \mathbf{W}^K_{i} \mathbf{X}
\end{equation}

\begin{figure}
    \centering
    \includegraphics[width=197pt]{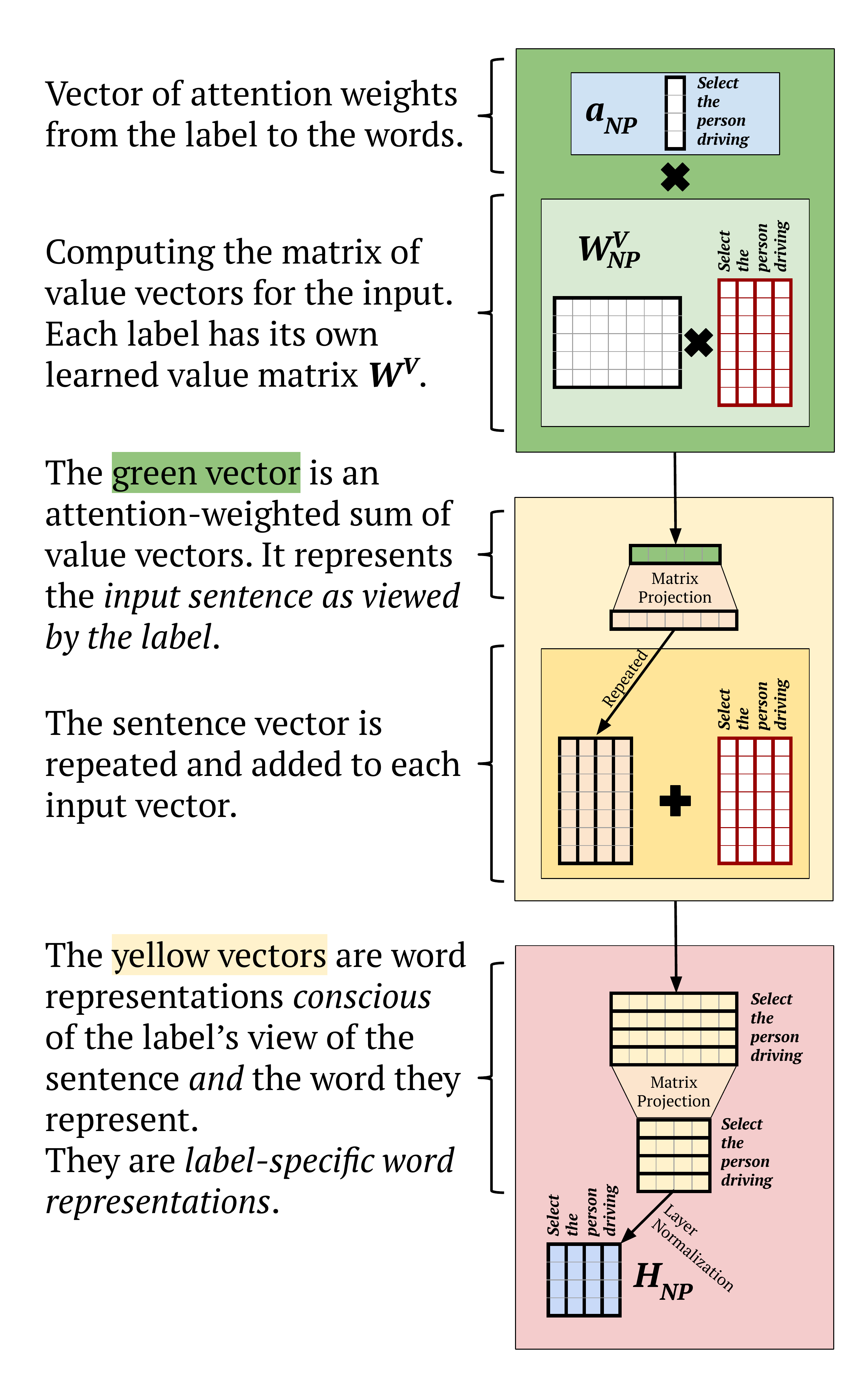}
    \caption{The Value vector computations in our proposed Label Attention Layer.}
    \label{lal5}
\end{figure}

Each attention head in our Label Attention layer has an attention \textit{vector}, instead of an attention \textit{matrix} as in self-attention. Consequently, we do not obtain a \textit{matrix} of vectors, but a \textit{single} vector that contains head-specific context information. This \textit{context} vector corresponds to the green vector in Figure \ref{lal5}. We compute the context vector $\mathbf{c}_i$ of head $i$ as follows:

\begin{equation}
    \mathbf{c}_i = \mathbf{a}_i * \mathbf{V}_i
\end{equation}

\noindent where $\mathbf{a}_i$ is the vector of attention weights in Equation \ref{eq1}, and $\mathbf{V}_i$ is the matrix of value vectors. Given a learned head-specific value matrix $\mathbf{W}^V_{i}$, we compute $\mathbf{V}_i$ as:

\begin{equation}
    \mathbf{V}_i = \mathbf{W}^V_{i} \mathbf{X}
\end{equation}

\begin{figure*}
    \centering
    \includegraphics[width=420pt]{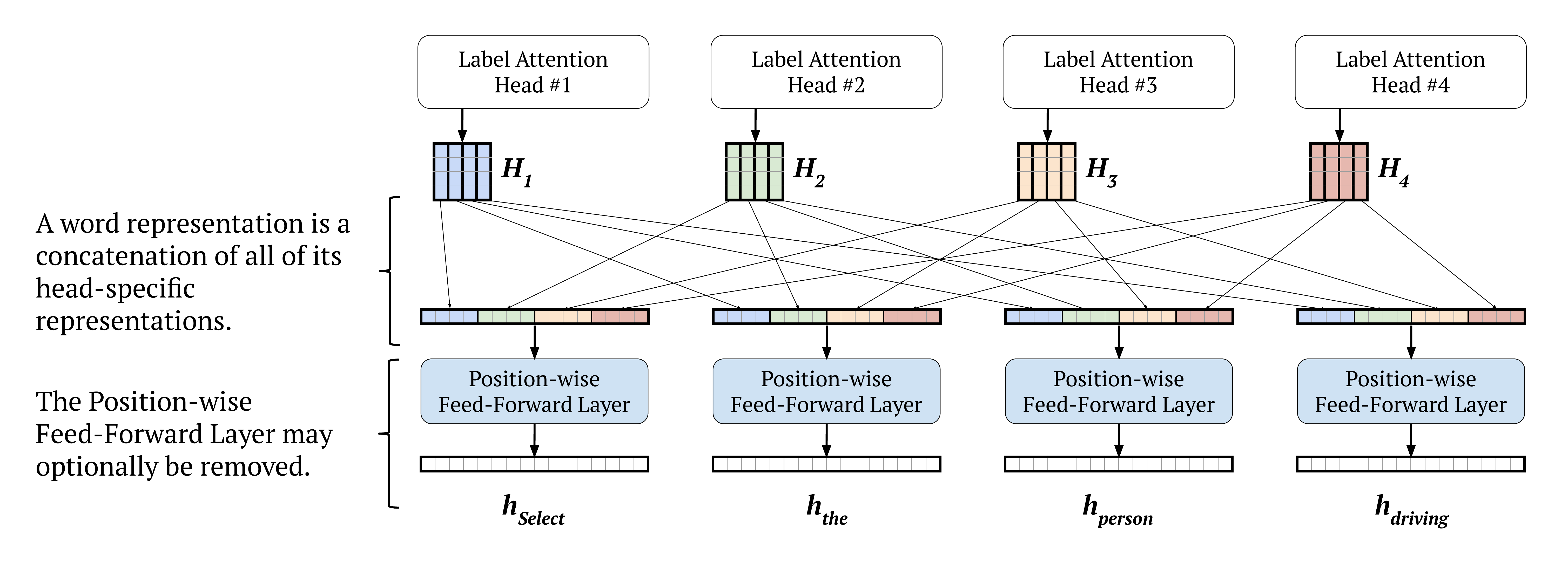}
    \caption{Redistribution of the head-specific word representations to form word vectors by concatenation. We use different colors for each label attention head. The colors show where the head outputs go in the word representations. We do not use colors for the vectors resulting from the position-wise feed-forward layer, as the head-specific information moved.}
    \label{lal6}
\end{figure*}

The context vector gets added to each individual input vector – making for one residual connection per head, rather one for all heads, as in the yellow box in Figure \ref{lal5}. We project the resulting matrix of word vectors to a lower dimension before normalizing. We then distribute the vectors computed by each label attention head, as shown in Figure \ref{lal6}.

We chose to assign as many attention heads to the Label Attention Layer as there are classification labels. As parsing labels (syntactic categories) are related, we did not apply an orthogonality loss to force the heads to learn separate information. We therefore expect an overlap when we match labels to heads. The values from each head are identifiable within the final word representation, as shown in the color-coded vectors in Figure \ref{lal6}.

The activation functions of the position-wise feed-forward layer make it difficult to follow the path of the contributions. Therefore we can remove the position-wise feed-forward layer, and compute the contributions from each label. We provide an example in Figure \ref{lal3}, where the contributions are computed using normalization and averaging. In this case, we are computing the contributions of each head to the span vector. The span representation for ``\textit{the person}'' is computed following the method of \citet{gaddy2018s} and \citet{kitaev2018constituency}. However, forward and backward representations are not formed by splitting the entire word vector at the middle, but rather by splitting each head-specific word vector at the middle.

\begin{figure}
    \centering
    \includegraphics[width=197pt]{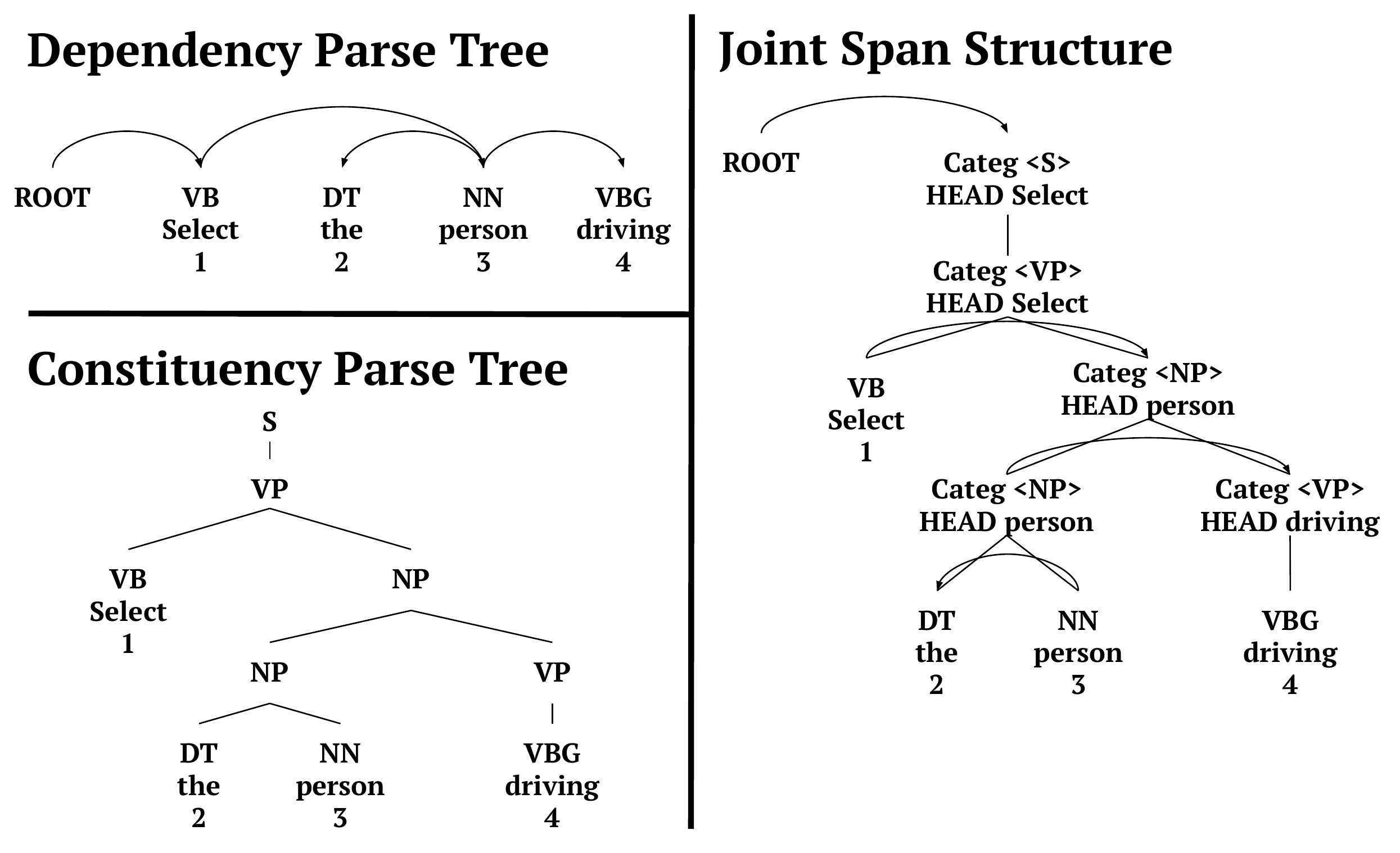}
    \caption{Parsing representations of the example sentence in Figure \ref{lal4}.}
    \label{lal_trees}
\end{figure}

In the example in Figure \ref{lal3}, we show averaging as one way of computing contributions, other functions, such as softmax, can be used. Another way of interpreting predictions is to look at the head-to-word attention distributions, which are the output vectors in the computation in Figure \ref{lal4}.

\section{Syntactic Parsing Model}
\label{section_parser}

\subsection{Encoder}
Our parser is an encoder-decoder model. The encoder has self-attention layers \cite{vaswani2017attention}, preceding the Label Attention Layer. We follow the attention partition of \citet{kitaev2018constituency}, who show that separating content embeddings from position ones improves performance.

Sentences are pre-processed following \citet{zhou2019head}. Trees are represented using a simplified Head-driven Phrase Structure Grammar (HPSG) \cite{pollard1994head}. In \citet{zhou2019head},   two kinds of span representations are proposed: the division span and the joint span. We choose the joint span representation as it is the best-performing one in their experiments.  Figure \ref{lal_trees} shows how the example sentence in Figure \ref{lal4} is represented.

The token representations for our model are a concatenation of content and position embeddings. The content embeddings are a sum of word and part-of-speech embeddings.

\subsection{Constituency Parsing}

\begin{figure*}
    \centering
    \includegraphics[width=420pt]{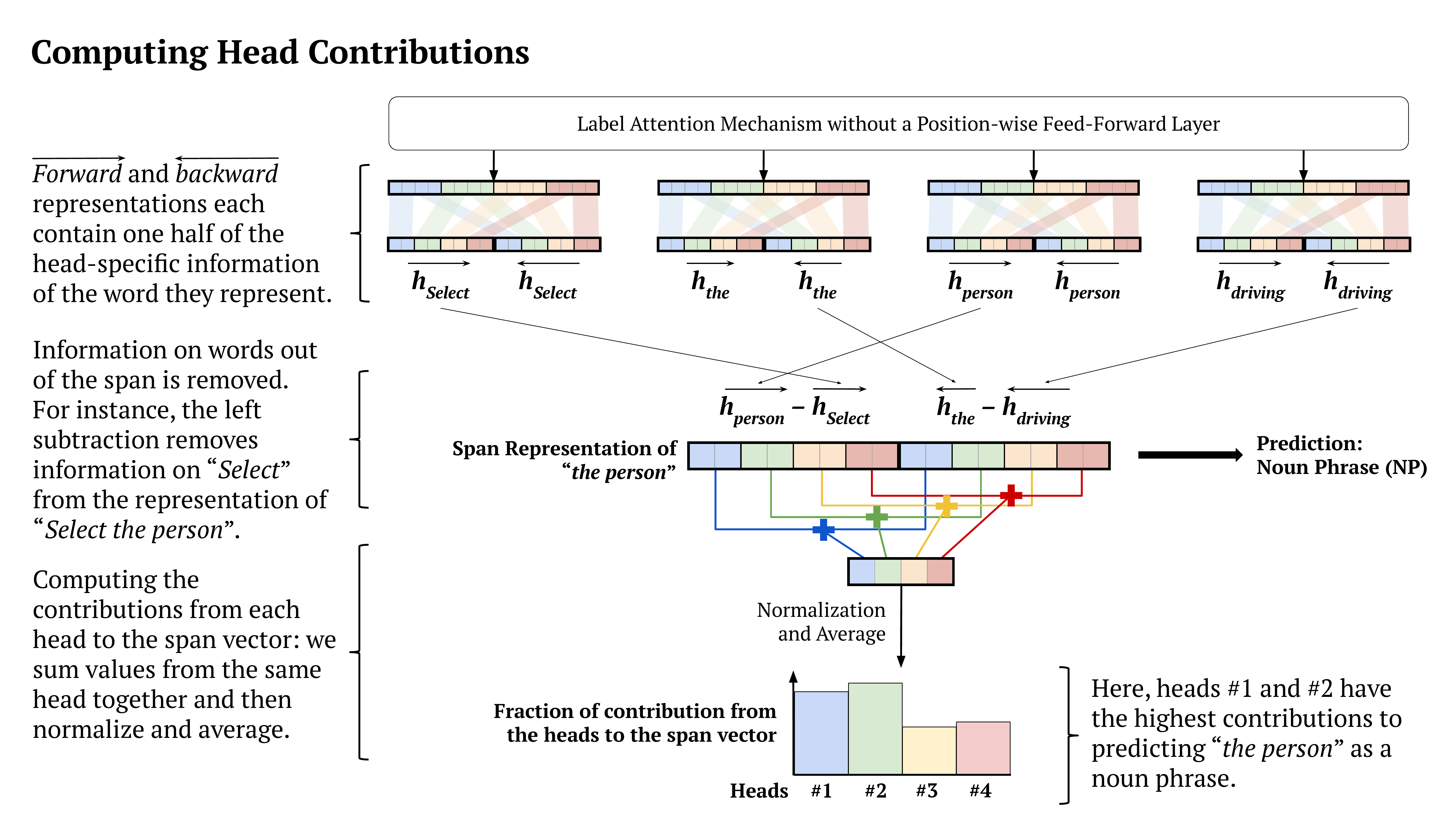}
    \caption{If we remove the position-wise feed-forward layer, we can compute the contributions from each label attention head to the span representation, and thus interpret head contributions. This illustrative example follows the label color scheme in Figure \ref{lal6}.}
    \label{lal3}
\end{figure*}

For constituency parsing, span representations follow the definition of \citet{gaddy2018s} and \citet{kitaev2018constituency}. For a span starting at the $i$-th word and ending at the $j$-th word, the corresponding span vector $s_{ij}$ is computed as:

\begin{equation}
    \mathbf{s_{ij}} = \left[\overrightarrow{\mathbf{h_j}} - \overrightarrow{\mathbf{h_{i-1}}}; \overleftarrow{\mathbf{h_{j+1}}} - \overleftarrow{\mathbf{h_i}}\right]
\end{equation}

\noindent where $\overleftarrow{\mathbf{h_i}}$ and $\overrightarrow{\mathbf{h_i}}$ are respectively the backward and forward representation of the $i$-th word obtained by splitting its representation in half. An example of a span representation is shown in the middle of Figure \ref{lal3}.

The score vector for the span is obtained by applying a one-layer feed-forward layer:

\begin{equation}
    \mathbf{S}(i,j) = \mathbf{W_2} \textup{ReLU} 
    ( \textup{LN} 
    ( \mathbf{W_1} \mathbf{s_{ij}}
    + \mathbf{b_1} ) ) 
    + \mathbf{b_2}
\end{equation}

\noindent where $\textup{LN}$ is Layer Normalization, and $\mathbf{W_1}$, $\mathbf{W_2}$, $\mathbf{b_1}$ and $\mathbf{b_2}$ are learned parameters. For the $l$-th syntactic category, the corresponding score $s(i,j,l)$ is then the $l$-th value in the $\mathbf{S}(i,j)$ vector.

Consequently, the score of a constituency parse tree $T$ is the sum of all of the scores of its spans and their syntactic categories:

\begin{equation}
    s(T) = \sum_{(i,j,l) \in T} s(i,j,l)
\end{equation}

We then use a CKY-style algorithm \cite{stern2017minimal, gaddy2018s} to find the highest scoring tree $\hat{T}$. The model is trained to find the correct parse tree $T^{*}$, such that for all trees $T$, the following margin constraint is satisfied:

\begin{equation}
    s(T^{*}) \geq s(T) + \Delta (T, T^{*})
\end{equation}

\noindent where $\Delta$ is the Hamming loss on labeled spans. The corresponding loss function is the hinge loss:

\begin{equation}
    L_{c} = \textup{max}\left(0, \textup{max}_{T}[s(T) + \Delta (T, T^{*})] - s(T^{*})\right)
\end{equation}

\subsection{Dependency Parsing}

We use the biaffine attention mechanism \cite{dozat2016deep} to compute a probability distribution for the dependency head of each word. The child-parent score $\alpha_{ij}$ for the $j$-th word to be the head of the $i$-th word is:

\begin{equation}
    \alpha_{ij} = \mathbf{h_i^{(d)}}^T \mathbf{W} \mathbf{h_j^{(h)}} + \mathbf{U}^T \mathbf{h_i^{(d)}} + \mathbf{V}^T \mathbf{h_j^{(h)}} + b
\end{equation}

\noindent where $\mathbf{h_i^{(d)}}$ is the dependent representation of the $i$-th word obtained by putting its representation $\mathbf{h_i}$ through a one-layer perceptron. Likewise, $\mathbf{h_j^{(h)}}$ is the head representation of the $j$-th word obtained by putting its representation $\mathbf{h_j}$ through a separate one-layer perceptron. The matrices $\mathbf{W}$, $\mathbf{U}$ and $\mathbf{V}$ are learned parameters.

The model trains on dependency parsing by minimizing the negative likelihood of the correct dependency tree. The loss function is cross-entropy:

\begin{equation}
    L_d = - \textup{log}\left(P\left(h_i|d_i\right)P\left(l_i|d_i, h_i\right)\right)
\end{equation}

\noindent where $h_i$ is the correct head for dependent $d_i$, $P\left(h_i|d_i\right)$ is the probability that $h_i$ is the head of $d_i$, and $P\left(l_i|d_i, h_i\right)$ is the probability of the correct dependency label $l_i$ for the child-parent pair $(d_i, h_i)$.

\subsection{Decoder}

The model jointly trains on constituency and dependency parsing by minimizing the sum of the constituency and dependency losses:

\begin{equation}
    L = L_c + L_d
\end{equation}

The decoder is a CKY-style \cite{kasami1966efficient, younger1967recognition, cocke1969programming, stern2017minimal} algorithm, modified by \citet{zhou2019head} to include dependency scores.

\section{Experiments}
\label{section_exp}

We evaluate our model on the English Penn Treebank (PTB) \cite{marcus1993building} and on the Chinese Treebank (CTB) \cite{xue2005penn}. We use the Stanford tagger \cite{toutanova2003feature} to predict part-of-speech tags and follow standard data splits.

Following standard practice, we use the \texttt{EVALB} algorithm \cite{sekine1997evalb} for constituency parsing, and report results without punctuation for dependency parsing.

\subsection{Setup}
In our English-language experiments, the Label Attention Layer has 112 heads: one per syntactic category. However, this is an experimental choice, as the model is not designed to have a one-on-one correspondence between attention heads and syntactic categories. The Chinese Treebank is a smaller dataset, and therefore we use 64 heads in Chinese-language experiments, even though the number of Chinese syntactic categories is much higher. For both languages, the query, key and value vectors, as well as the output vectors of each label attention head, have 128 dimensions, as determined through short parameter-tuning experiments. For the dependency and span scores, we use the same hyperparameters as \citet{zhou2019head}. We use the large cased pre-trained XLNet \cite{yang2019xlnet} as our embedding model for our English-language experiments, and a base pre-trained BERT \cite{devlin2018bert} for Chinese. 

We try English-language parsers with 2, 3, 4, 6, 8, 12 and 16 self-attention layers. Our parsers with 3 and 4 self-attention layers are tied in terms of F1 score, and sum of UAS and LAS scores. The results of our fine-tuning experiments are in the appendix. We decide to use 3 self-attention layers for all the following experiments, for lower computational complexity.

\subsection{Ablation Study}

As shown in Figure \ref{lal3}, we can compute the contributions from label attention heads only if there is no position-wise feed-forward layer. Residual dropout in self-attention applies to the aggregated outputs from all heads. In label attention, residual dropout applies separately to the output of each head, and therefore can cancel out parts of the head contributions. We investigate the impact of removing these two components from the LAL.

We show the results on the PTB dataset of our ablation study on Residual Dropout and Position-wise Feed-forward Layer in Table \ref{exp2}. We use the same residual dropout probability as \citet{zhou2019head}. When removing the position-wise feed-forward layer and keeping residual dropout, we observe only a slight decrease in overall performance, as shown in the second row. There is therefore no significant loss in performance in exchange for the interpretability of the attention heads.

\begin{table}[]
    \centering
     \small
    \begin{tabular}{|l|l|r|r|r|r|r|}
        \hline
        \bf PFL & \bf RD  & \bf Prec. & \bf Recall & \bf F1 & \bf UAS & \bf LAS \\ \hline
        Yes & Yes & 96.47 & 96.20 & 96.34 & 97.33 & \bf 96.29 \\
        No & Yes & 96.51 & 96.15 & 96.33 & 97.25 & 96.11 \\
        Yes & No & \bf 96.53 & \bf 96.24 & \bf 96.38 & \bf 97.42 & 96.26 \\
        No & No & 96.29 & 96.05 & 96.17 & 97.23 & 96.11 \\ \hline
    \end{tabular}
    \caption{Results on the PTB test set of the ablation study on the Position-wise Feed-forward Layer (\textbf{PFL}) and Residual Dropout (\textbf{RD}) of the Label Attention Layer.}
    \label{exp2}
\end{table}

\begin{table}[]
    \centering
     \small
    \begin{tabular}{|l|l|r|r|r|r|r|}
        \hline
        \bf QV & \bf Conc. & \bf Prec. & \bf Recall & \bf F1 & \bf UAS & \bf LAS \\ \hline
        Yes & Yes & \bf 96.53 & \bf 96.24 & \bf 96.38 & \bf 97.42 & \bf 96.26 \\
        No & Yes & 96.43 & 96.03 & 96.23 & 97.25 & 96.12 \\ 
        Yes & No & 96.30 & 96.10 & 96.20 & 97.23 & 96.15 \\ 
        No & No & 96.30 & 96.06 & 96.18 & 97.26 & 96.17 \\ 
        \hline
    \end{tabular}
    \caption{Results on the PTB test set of the ablation study on the Query Vectors (\textbf{QV}) and Concatenation (\textbf{Conc.}) parts of the Label Attention Layer.}
    \label{exp4}
\end{table}

\begin{figure}
    \centering
    \includegraphics[width=220pt]{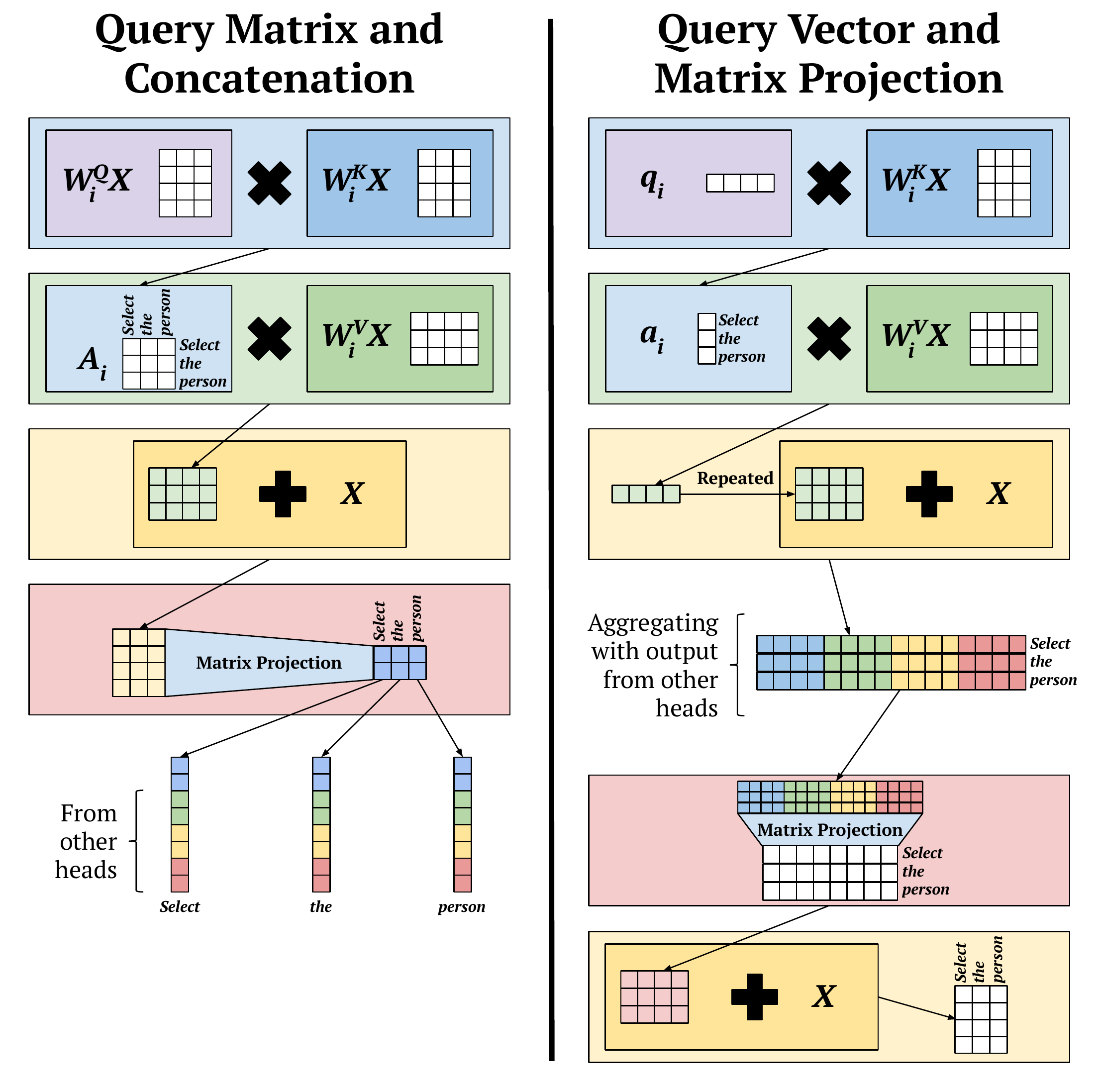}
    \caption{The two hybrid parser architectures for the ablation study on the Label Attention Layer's Query Vectors and Concatenation.}
    \label{lalc2}
\end{figure}

We observe an increase in performance when removing residual dropout only. This suggests that all head contributions are important for performance, and that we were likely over-regularizing.

Finally, removing both position-wise feed-forward layer and residual dropout brings about a noticeable decrease in performance. We continue our experiments without residual dropout.

\subsection{Comparison with Self-Attention}

The two main architecture novelties of our proposed Label Attention Layer are the learned Query Vectors that represent labels and replace the Query Matrix in self-attention, and the Concatenation of the outputs of each attention head that replaces the Matrix Projection in self-attention.

In this subsection, we evaluate whether our proposed architecture novelties bring about performance improvements. To this end, we establish an ablation study to compare Label Attention with Self-Attention. We propose three additional model architectures based on our best parser: all models have 3 self-attention layers and a modified Label Attention Layer with 112 attention heads. The three modified Label Attention Layers are as follows: \textbf{(1) Ablation of Query Vectors:} the first model (left of Figure \ref{lalc2}) has a Query Matrix like self-attention, and concatenates attention head outputs like Label Attention. \textbf{(2) Ablation of Concatenation:} the second model (right of Figure \ref{lalc2}) has a Query Vector like Label Attention, and applies matrix projection to all head outputs like self-attention. \textbf{(3) Ablation of Query Vectors and Concatenation:} the third model (right of Figure \ref{lalc}) has a 112-head self-attention layer.

The results of our experiments are  in Table \ref{exp4}. The second row shows that, even though query matrices employ more parameters and computation than query vectors, replacing query vectors by query matrices decreases performance. There is a similar decrease in performance when removing concatenation as well, as shown in the last row. This suggests that our Label Attention Layer learns meaningful representations in its query vectors, and that head-to-word attention distributions are more helpful to performance than query matrices and word-to-word attention distributions.

In self-attention, the output vector is a matrix projection of the concatenation of head outputs. In Label Attention, the head outputs do not interact through matrix projection, but are concatenated. The third and fourth rows of Table \ref{exp4} show that there is a significant decrease in performance when replacing concatenation with the matrix projection. This decrease suggests that the model benefits from having one residual connection per attention head, rather than one for all attention heads, and from separating head-specific information in word representations. In particular, the last row shows that replacing our LAL with a self-attention layer with an equal number of attention heads decreases performance: the difference between the performance of the first row and the last row is due to the Label Attention Layer's architecture novelties.

\subsection{English and Chinese Results}

\begin{table}[]
    \centering
    \small
    \resizebox{\columnwidth}{!}{
    \begin{tabular}{|p{3cm}|l|l|l|l|l|l|}
        \hline
        \multirow{2}{*}{\bf Model} & \multicolumn{3}{c|}{\bf English} & \multicolumn{3}{c|}{\bf Chinese} \\ \cline{2-7}
        & \bf LR & \bf LP & \bf F1 & \bf LR & \bf LP & \bf F1 \\ \hline
        \citet{shen-etal-2018-straight} & 92.0 & 91.7 & 91.8 & 86.6 & 86.4 & 86.5 \\
        \citet{fried-klein-2018-policy} & - & - & 92.2 & - & - & 87.0 \\
        \citet{teng-zhang-2018-two} & 92.2 & 92.5 & 92.4 & 86.6 & 88.0 & 87.3 \\
        \citet{vaswani2017attention} & - & - & 92.7 & - & - & - \\
        \citet{dyer2016recurrent} & - & - & 93.3 & - & - & 84.6 \\
        \citet{kuncoro2017recurrent} & - & - & 93.6 & - & - & -  \\
        \citet{charniak2016parsing} & - & - & 93.8 & - & - & - \\
        \citet{liu-zhang-2017-shift} &  91.3 & 92.1 & 91.7 & 85.9 & 85.2 & 85.5 \\
        \citet{liu2017order} & - & - & 94.2 & - & - & 86.1 \\
        \citet{suzuki2018empirical} & - & - & 94.32 & - & - & -  \\
        \citet{takase2018direct} & - & - & 94.47& - & - & -  \\
        \citet{fried2017improving} & - & - & 94.66 & - & - & - \\
        \citet{kitaev2018constituency} & 94.85 & 95.40 & 95.13 & - & - & - \\
        \citet{kitaev2018multilingual} & 95.51 & 96.03 & 95.77 & 91.55 & 91.96 & 91.75 \\
        \citet{zhou2019head} (BERT) & 95.70 & 95.98 & 95.84 &\bf 92.03 & 92.33 & 92.18 \\
        \citet{zhou2019head} (XLNet) & 96.21 & 96.46 & 96.33 & - & - & - \\
        \hline
        Our work & \bf 96.24 & \bf 96.53 & \bf 96.38 & 91.85 & \bf 93.45 & \bf 92.64 \\ \hline
    \end{tabular}}
    \caption{Constituency Parsing on PTB \& CTB test sets.}
    \label{hpsg_comparison_const}
\end{table}

Our best-performing English-language parser does not have residual dropout, but has a position-wise feed-forward layer. We train Chinese-language parsers using the same configuration. The Chinese Treebank has two data splits for the training, development and testing sets: one for Constituency \cite{liu-zhang-2017-shift} and one for Dependency parsing \cite{zhang-clark-2008-tale}. 

Finally, we compare our results with the state of the art in constituency and dependency parsing in both English and Chinese. We show our Constituency Parsing results in Table \ref{hpsg_comparison_const}, and our Dependency Parsing results in Table \ref{hpsg_comparison_dep}. Our LAL parser establishes new state-of-the-art results in both languages, improving significantly in dependency parsing.

\begin{table}[]
    \centering
    \resizebox{\columnwidth}{!}{
    \begin{tabular}{|p{5cm}|l|l|l|l|}
        \hline
        \multirow{2}{*}{\bf Model} & \multicolumn{2}{c|}{\bf English} & \multicolumn{2}{c|}{\bf Chinese} \\ \cline{2-5}
        & \bf UAS & \bf LAS & \bf UAS & \bf LAS \\ \hline
        \citet{kuncoro-etal-2016-distilling} & 94.26 & 92.06 &  88.87 & 87.30 \\
        \citet{li-etal-2018-seq2seq} & 94.11 & 92.08 & 88.78 & 86.23 \\
        \citet{ma-hovy-2017-neural} & 94.88 & 92.98 & 89.05 & 87.74 \\
        \citet{dozat2016deep} &  95.74 & 94.08 & 89.30 & 88.23 \\
        \citet{choe-charniak-2016-parsing} & 95.9 & 94.1 & - & - \\
        \citet{ma-etal-2018-stack} & 95.87 & 94.19 & 90.59 & \bf 89.29 \\
        \citet{ji2019graph} & 95.97 & 94.31 & - & - \\
        \citet{fernandez2019left} &  96.04 & 94.43 & - & - \\
        \citet{kuncoro2017recurrent} & 95.8 & 94.6 & - & - \\
        \citet{clark2018semi} &  96.61 & 95.02 & - & - \\
        \citet{wang-etal-2018-improved} & 96.35  & 95.25 & - & - \\
        \citet{zhou2019head} (BERT) & 97.00 & 95.43 & 91.21 & 89.15 \\
        \citet{zhou2019head} (XLNet) & 97.20 & 95.72 & - & - \\
        \hline
        Our work & \bf 97.42 & \bf 96.26 & \bf 94.56 & 89.28 \\ \hline
    \end{tabular}}
    \caption{Dependency Parsing on PTB \& CTB test sets.}
    \label{hpsg_comparison_dep}
\end{table}

\subsection{Interpreting Head Contributions}

We follow the method in Figure \ref{lal3} to identify which attention heads contribute to predictions. We collect the span vectors from the Penn Treebank test set, and we use our LAL parser with no position-wise feed-forward layer for predictions.

\begin{figure}
    \centering
    \includegraphics[width=215pt]{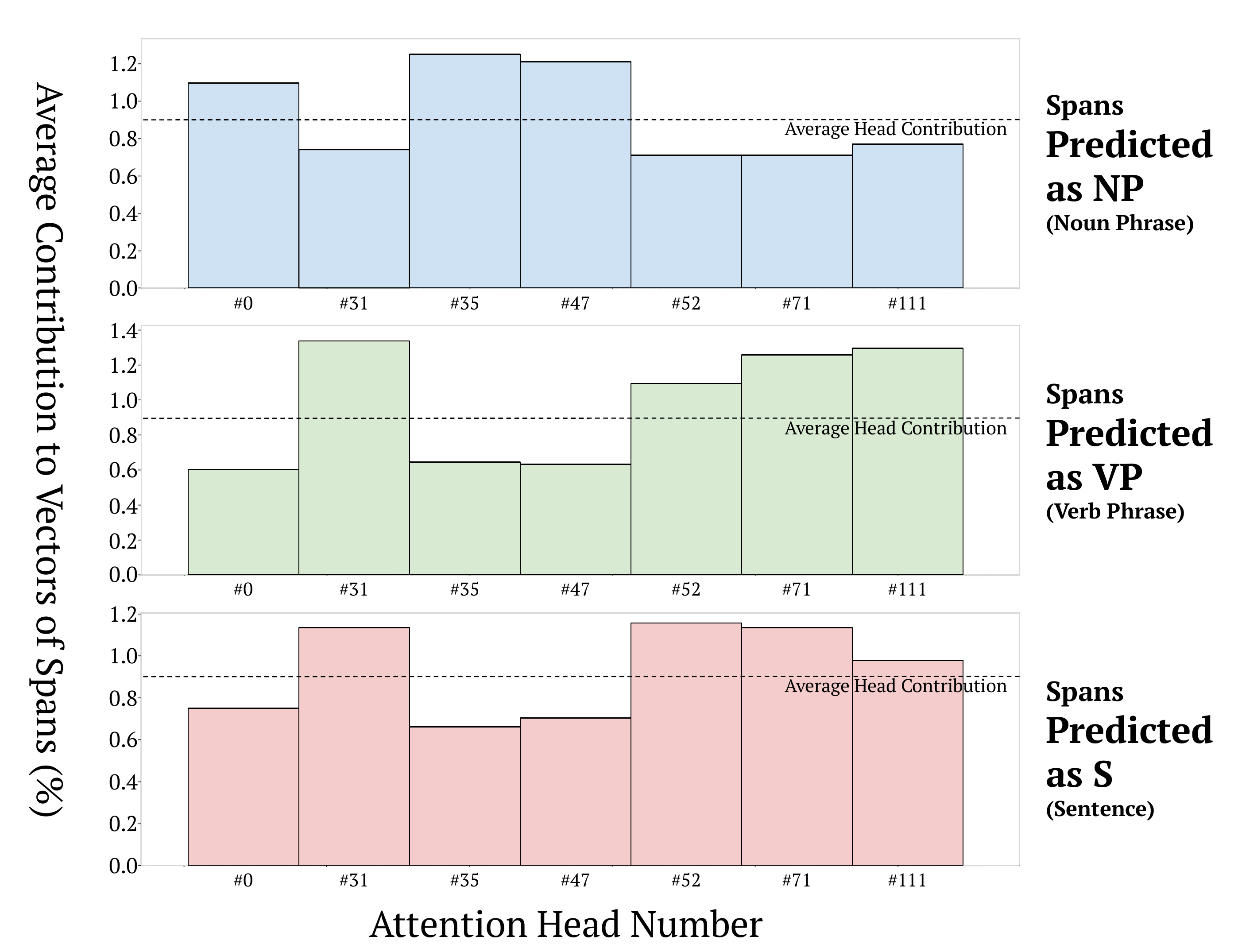}
    \caption{Average contribution of select heads to span vectors with different predicted syntactic categories.}
    \label{lalb}
\end{figure}

Figure \ref{lalb} displays the bar charts for the three most common syntactic categories: 
Noun Phrases (NP), Verb Phrases (VP) and Sentences (S). 
We notice several heads explain each predicted category.

We collect statistics about the top-contributing heads for each predicted category. Out of the NP spans, 44.9\% get their top contribution from head 35, 13.0\% from head 47, and 7.3\% from head 0. The top-contributing heads for VP spans are heads 31 (61.1\%), 111 (13.2\%), and 71 (7.5\%). As for S spans, the top-contributing heads are 52 (48.6\%), 31 (22.8\%), 35 (6.9\%), and 111 (5.2\%). We see that S spans share top-contributing heads with VP spans (heads 31 and 111), and NP spans (head 35). The similarities reflect the relations between the syntactic categories. In this case, our Label Attention Layer learned the rule S $\rightarrow$ NP VP.

Moreover, the top-contributing heads for PP spans are 35 (29.6\%), 31 (26.7\%), 111 (10.3\%), and 47 (9.4\%): they are equally split between NP spans (heads 35 and 47) and VP spans (heads 31 and 111). Here, the LAL has learned that both verb and noun phrases can contain preposition phrases.

We see that head 52 is unique to S spans. Actually, 64.7\% of spans with head 52 as the highest contribution are S spans. Therefore our model has learned to represent the label S using head 52.

All of the aforementioned heads are represented in Figure \ref{lalb}. We see that heads that have low contributions for NP spans, peak in contribution for VP spans (heads 31, 71 and 111), and vice-versa (heads 0, 35 and 47). Moreover, NP spans do not share any top-contributing head with VP spans. This shows that our parser has also learned the differences between dissimilar syntactic categories.

\subsection{Error Analysis}
{\bf Head-to-Word Attention. } We analyze prediction errors from the PTB test set. One example is the span ``\textit{Fed Ready to Inject Big Funds}'', predicted as NP but labelled as S. We trace back the attention weights for each word, and find that, out of the 9 top-contributing heads, only 2 focus their attention on the root verb of the sentence (\textit{Inject}), while 4 focus on a noun (\textit{Funds}), resulting in a noun phrase prediction. We notice similar patterns in other wrongly predicted spans, suggesting that forcing the attention distribution to focus on a relevant word might correct these errors.

{\bf Top-Contributing Heads. } We analyze wrongly predicted spans by their true category. Out of the 53 spans labelled as NP but not predicted as such, we still see the top-contributing head for 36 of them is either head 35 or 47, both top-contributing heads of spans predicted as NP. Likewise, for the 193 spans labelled as S but not predicted as such, the top-contributing head of 141 of them is one of the four top-contributing heads for spans predicted as S. This suggests that a stronger prediction link to the label attention heads, through a loss function for instance, may increase the performance.

\section{Related Work}
\label{section_rel}

Since their introduction in Machine Translation, attention mechanisms \cite{bahdanau2014neural, luong2015effective} have been extended to other tasks, such as text classification \cite{yang2016hierarchical}, natural language inference \cite{chen2016enhancing} and language modeling \cite{salton2017attentive}.

Self-attention and transformer architectures \cite{vaswani2017attention} are now the state of the art in language understanding \cite{devlin2018bert, yang2019xlnet}, extractive summarization \cite{DBLP:journals/corr/abs-1903-10318}, semantic role labeling \cite{strubell2018linguistically} and machine translation for low-resource languages \cite{rikters2018impact, rikters2018training}.


While attention mechanisms can provide explanations for model predictions, \citet{serrano2019attention} challenge that assumption and find that attention weights only noisily predict overall importance with regard to the model. \citet{jain2019attention} find that attention distributions rarely correlate with feature importance weights. However, \citet{wiegreffe2019attention} show through alternative tests that prior work does not discredit the usefulness of attention for interpretability.

\citet{xiao2019label} introduce the Label-Specific Attention Network (LSAN) for multi-label document classification. They use label descriptions to compute attention scores for words, and follow the self-attention of \citet{lin2017structured}. \citet{cui2019hierarchically} introduce a Label Attention Inference Layer for sequence labeling, which uses the self-attention of \citet{vaswani2017attention}. In this case, the key and value vectors are learned label embeddings, and the query vectors are hidden vectors obtained from a Bi-LSTM encoder. Our work is unrelated to these two papers, as they were published towards the end of our project.

\section{Conclusions}
\label{section_conc}

In this paper, we introduce a new form of self-attention: the Label Attention Layer. In our proposed architecture, attention heads represent labels. We incorporate our Label Attention Layer into the HPSG parser \cite{zhou2019head} and obtain new state-of-the-art results on the Penn Treebank and Chinese Treebank. In English, our results show 96.38 F1 for constituency parsing, and 97.42 UAS and 96.26 LAS for dependency parsing. In Chinese, our model achieves 92.64 F1, 94.56 UAS and 89.28 LAS. 

We perform ablation studies that show the Query Vector learned by our Label Attention Layer outperform the self-attention Query Matrix.  Since we have only one learned vector as query, rather than a matrix, we can significantly reduce the number of parameters per attention head. Finally, our Label Attention heads learn the relations between the syntactic categories, as we show by computing contributions from each attention head to span vectors. We show how the heads also help to analyze prediction errors, and suggest methods to correct them.

\section*{Acknowledgements}

We thank the anonymous reviewers for their helpful and detailed comments.

\bibliography{anthology,acl2020}
\bibliographystyle{acl_natbib}

\appendix

\section{Additional Experiment Results}

We report experiment results for hyperparameter tuning based on the number of self-attention layers in Table \ref{layers_appendix}.

\begin{table*}[]
    \centering
    \begin{tabular}{|r|r|r|r|r|r|}
        \hline
        \bf Self-Attention Layers & \bf Precision & \bf Recall & \bf F1 & \bf UAS & \bf LAS \\ \hline
        2  & 96.23 & 96.03 & 96.13 & 97.16 & 96.09 \\
        3  & 96.47 & \bf 96.20 & \bf 96.34 & 97.33 & \bf 96.29 \\
        4  & \bf 96.52 & 96.15 & \bf 96.34 & \bf 97.39 & 96.23 \\
        6  & 96.48 & 96.09 & 96.29 & 97.30 & 96.16 \\
        8  & 96.43 & 96.09 & 96.26 & 97.33 & 96.15 \\
        12 & 96.27 & 96.06 & 96.16 & 97.24 & 96.14 \\
        16 & 96.38 & 96.02 & 96.20 & 97.32 & 96.11 \\ \hline
    \end{tabular}
    \caption{Performance on the Penn Treebank test set of our LAL parser according to the number of self-attention layers. All parsers here include the Position-wise Feed-forward Layer and Residual Dropout.}
    \label{layers_appendix}
\end{table*}


\end{document}